\documentclass{article}

\usepackage[nonatbib, final]{neurips_2024_ml4ps}
\usepackage[utf8]{inputenc} % allow utf-8 input
\usepackage[T1]{fontenc}    % use 8-bit T1 fonts
\usepackage{hyperref}       % hyperlinks
\usepackage{url}            % simple URL typesetting
\usepackage{booktabs}       % professional-quality tables
\usepackage{amsfonts}       % blackboard math symbols
\usepackage{nicefrac}       % compact symbols for 1/2, etc.
\usepackage{microtype}      % microtypography
\usepackage[pdftex]{graphicx}
\usepackage{amsmath}
\usepackage{xcolor}
\usepackage{colortbl}
\usepackage{graphicx}
\usepackage{float}
\usepackage{ragged2e}
\usepackage{appendix}

\title{\textit{DYffCast}: Regional Precipitation Nowcasting Using IMERG Satellite Data. A case study over South America}

\author{Daniel Seal\textsuperscript{1}, Rossella Arcucci\textsuperscript{1,2}, Salva Rühling-Cachay\textsuperscript{3}, César Quilodrán-Casas\textsuperscript{1,4,5}\\
\textsuperscript{1}Department of Earth Science and Engineering, Imperial College London\\
\textsuperscript{2}Data Science Institute, Imperial College London\\
\textsuperscript{3}Department of Computer Science and Engineering, University of California San Diego\\
\textsuperscript{4}Grantham Institute for Climate Change and the Environment\\
\textsuperscript{5}National Center for Artificial Intelligence CENIA, Chile\\
}

\begin{document}

\maketitle

\begin{abstract}
Climate change is increasing the frequency of extreme precipitation events, making weather disasters such as flooding and landslides more likely. The ability to accurately nowcast precipitation is therefore becoming more critical for safeguarding society by providing immediate, accurate information to decision makers. Motivated by the recent success of generative models at precipitation nowcasting, this paper: extends the DYffusion framework to this task and evaluates its performance at forecasting IMERG satellite precipitation data up to a 4-hour horizon; modifies the DYffusion framework to improve its ability to model rainfall data; and introduces a novel loss function that combines MSE, MAE and the LPIPS perceptual score. In a quantitative evaluation of forecasts up to a 4-hour horizon, the modified DYffusion framework trained with the novel loss outperforms four competitor models. It has the highest CSI scores for weak, moderate, and heavy rain thresholds and retains an LPIPS score $<$ 0.2 for the entire roll-out, degrading the least as lead-time increases. The proposed nowcasting model demonstrates visually stable and sharp forecasts up to a 2-hour horizon on a heavy rain case study\footnote{Code is available at \url{https://github.com/Dseal95/DYffcast}}.
\end{abstract}

\section{Introduction}
The ability to accurately nowcast (predict up to 6 hours in advance \cite{wang2017}) precipitation, is a vital task that can help mitigate hazards such as flooding or landslides \cite{landslides}. Globally, heavy precipitation events are becoming more common due to climate change \cite{Donat2016, extremerain1, extremerain2, robinson2021, extremerain3}, causing large human displacement, human fatalities \cite{ocha2024latin}, and economic cost. Accurate precipitation forecasts are therefore becoming more instrumental in providing critical information required for weather-dependent decision making that can safeguard society. The absence of effective precipitation models especially impacts countries that lack access to freely available ground-based radar systems and experience vulnerability in resource management, agriculture, disaster preparedness, and climate adaptation.

This work evaluates the applicability of a novel probabilistic, spatio-temporal forecasting framework, DYffusion \cite{dyffusion}, to precipitation nowcasting. This research is motivated by the recent success of generative modelling at this task and, DYffusion's skilful forecasts over long horizons on Navier-Stokes flows and sea surface temperatures (SST); two systems that exhibit complex dynamics comparable to that of precipitation. In this study, the focus is on how well DYffusion can forecast up to a 4-hour horizon, how well these forecasts compare to deterministic and statistical baselines, and what modifications are required to apply DYffusion to precipitation nowcasting.

The contributions to the field of precipitation nowcasting can be summarised as follows:

\begin{enumerate}
\item Evaluation of a novel modelling framework, alternative to diffusion-based or Generative Adversarial Network (GAN)-based models, at precipitation nowcasting.
\item A focus on countries lacking freely available ground-based radar and most impacted by global warming.
\item Introduction of a novel loss function that demonstrates how guiding a pixel-wise distance-based loss with a perceptual loss improves a model's ability to learn small-scale features in imbalanced datasets such as precipitation.
\end{enumerate}

\section{Methods}
\subsection{Data}
This research uses the publicly available satellite precipitation product, IMERG. In particular, it uses the latest version (V07B \cite{imergv7}) of the half-hourly, Early Run product with a $0.1^{\text{o}}$ spatial resolution and precipitation rates in $\text{mm} \cdot \text{h}^{-1}$. The dataset is created from four 128$\times$128 grid boxes that cover the South American countries: Colombia, Ecuador and $\text{Per}\text{\'{u}}$. These are regions that are susceptible to increased flooding and have limited availability of weather radar\footnote{According to the World Meteorological Organisation's (WMO) radar database, South America has roughly $1$/$4$ of North America's or Europe's radar \cite{wmo_radar_2019}.}. The four 128$\times$128 grid boxes are stacked into dimensions: (N, S, C, H, W) where N is the number of samples, S is the sequence length (input and target) and C is the number of channels (1 for IMERG). The data is preprocessed to $[0, 1]$ using linear normalisation, $\text{log}(1+X)$ transform
and min-max normalisation. For this research, a 4-hour horizon or 8$\times$1$\times$128$\times$128 images, was chosen to match the latency of the Early-Run IMERG product and avoid any discontinuity of real-time data.

\subsection{Modelling}
\subsubsection{DYffusion Framework}
\begin{figure}
  \centering
  \includegraphics[width=0.7\linewidth]{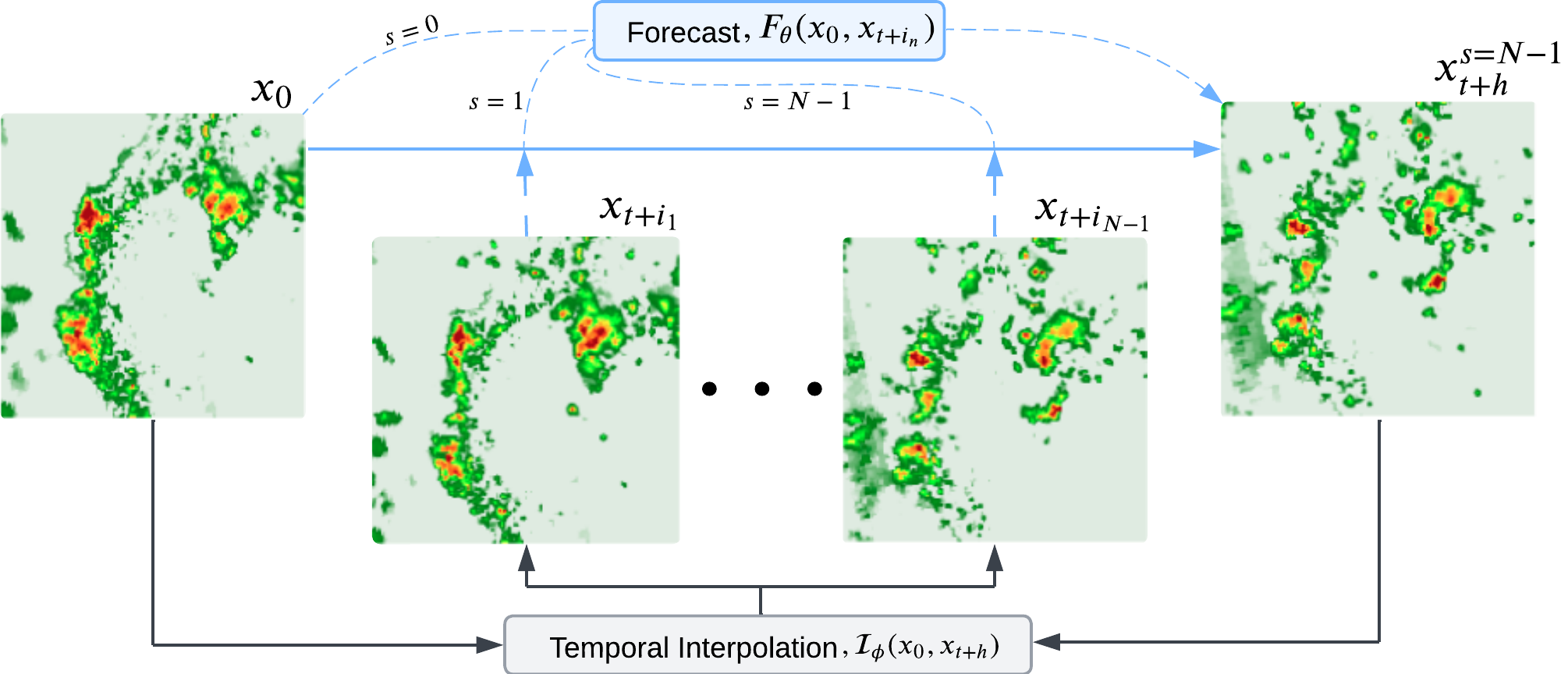}
  \caption{An illustration of DYffusion sampling: DYffusion forecasts a sequence of $N$ (or $h$) timesteps $x_{t+i_{1}}$ , $x_{t+i_{2}}$ , ... , $x_{t+h}$ given the initial condition $x_{0}$. $x_{t+h}$ is forecasted at each step using $x_{0}$ and the nearest intermediate value $x_{t+i_{n}}$.}
  \label{fig:dyffusion-method}
\end{figure}

The DYffusion framework is outlined in Figure \ref{fig:dyffusion-method}. It consists of an Interpolator network, $\mathrm{I}_{\phi}$ and a Forecastor network, $F_{\theta}$. 

\textbf{Training}. $\mathrm{I}_{\phi}$ is trained to interpolate any timestep of the sequence $x_{t+i_{n}}$, given the initial and final states, $x_{0}$ and $x_{t+h}$, respectively. $F_{\theta}$ is trained to forecast the horizon $x_{t+h}$ given the initial condition $x_{0}$ and an intermediate state $x_{t+i_{n}}$. This two-stage training has a constant memory footprint as a function of $h$, only requiring $x_{0}$ and $x_{t+h}$ (plus $x_{t+i_{n}}$ during the Interpolator training).

\textbf{Inference}. Given an initial condition, $x_{0}$, the entire sequence $x = [x_{t+i_{n}}]_{n=1}^{h}$ is generated over $N$ (or $h$) auto-regressive steps. At step $s$ = $0$ ($n$ = $s$ + 1), $F_{\theta}$ predicts the horizon $x_{t+h}^{s=0}$ using only $x_{0}$. $\mathrm{I}_{\phi}$ then interpolates $x_{t+i_{1}}$ using $x_{0}$ and $x_{t+h}^{s=0}$. At the next step ($s$ = $1$), $F_{\theta}$ predicts the horizon again $x_{t+h}^{s=1}$, using $x_{0}$ and $x_{t+i_{1}}$. $x_{t+h}^{s=1}$ and $x_{0}$ are then used to interpolate, $x_{t+i_{2}}$. This process is repeated until $F_{\theta}$ predicts $x_{t+h}^{s=N-1}$ ($N$ times), using the nearest intermediate value, $x_{t+i_{N-1}}$. This iterative process is analogous to the iterative denoising in standard diffusion models and underpins the time-coupled, DYnamics informed nature of the DYffusion framework. The final forecast is the mean of an ensemble of $X$ members.

The probabilistic nature of DYffusion occurs through Monte Carlo dropout. $\mathrm{I}_{\phi}$ is trained with dropout rates that are enabled during the training (and inference) of $F_{\theta}$. Enabling dropout during training is analogous to $\mathrm{I}_{\phi}$ interpolating $x_{t+i_{n}}$ by sampling it from the conditional probability distribution $P(x_{t+i_{n}} | x_{0}, x_{t+h})$.

\subsubsection{DYffusion Key Modifications}\label{sec:keymods}
For the IMERG dataset, the SST U-Net backbone from DYffusion and parts of the training framework were modified to model the rainfall data. The following key modifications were made:

\begin{enumerate}
\item \textbf{Loss Function.} Replaced the L1 loss with a novel loss function (referred to as LCB), combining the Learned Perceptual Image Patch Similarity (LPIPS)\cite{lpips} score  with the class-weighted Mean Squared Error (MSE) and Mean Absolute Error (MAE) loss proposed in \cite{trajgru} (referred to as CBLoss in \cite{atmosandradar}). LCB was constructed as follows:

\begin{equation}
\text{LCB} = (1-\alpha) \cdot \text{LPIPS} + \alpha \cdot \text{CBLoss}(\beta)
\end{equation}

Based on experimental results during Interpolator training, $\alpha$ was set to 0.6, optimising for MSE, LPIPS and Critical Success Index (CSI) at varying thresholds. This value slightly biases the \text{CBLoss} towards spatial accuracy, while the perceptual component prevents large-scale features from dominating and improves image detail. Lower $\alpha$ values ($<$ 0.6) led to background artefacts and magnitude errors. For the CBLoss, the $\beta$ scaling term was set to 1.0 to equally weight the MSE and MAE components, and the rainfall classes (in $\text{mm} \cdot \text{h}^{-1}$) and their associated weights were:
\begin{equation}
w : w(x) = \left\{
\begin{array}{lll}
1,\, x \leq 0.5 & 2,\, 0.5 < x \leq 2 & 5,\, 2 < x \leq 6 \\[1ex]
10,\, 6 < x \leq 10 & 20,\, 10 < x \leq 18 & 30,\, 18 < x \leq 30 \\[1ex]
50,\, x > 30
\end{array}
\right.
\end{equation}

\item \textbf{Forecastor Training.} Removed any exposure to the horizon $x_{t+h}$ during Forecastor training to better emulate the sampling process. Referring to Algorithm 1, Stage 2, Step 2 in the original DYffusion paper, the intermediate values $x_{t+i_{n}}$ are now interpolated using the initial condition $x_{0}$ and an initial forecast $\hat{x}_{t+h}^{inital}$ instead of the target $x_{t+h}$. This initial forecast is generated using only the initial conditions ($F_{\theta}(x_{0}, x_{0}) = \hat{x}_{t+h}^{initial}$) to directly replicate the sampling procedure. A separate loss term for the initial forecast is added to the existing loss in Stage 2, Step 3 of Algorithm 1 in the original DYffusion paper. The updated loss function (including the one-step-ahead loss) becomes:
\begin{equation}
\begin{split}
\text{\textbf{L}oss} &= \alpha\textbf{L}_{initial}+(1-\alpha)[\lambda_{1}\textbf{L}_{forecast} +\lambda_{2}\textbf{L}_{one-step-ahead}]
\end{split}
\end{equation}

Where \textbf{L} represents the loss function and $\alpha$ is used to balance the different terms. For training, a linear schedule is used to calculate $\alpha$, starting at $\alpha = 1$ (decaying to 0 over 20 epochs) to place more emphasis on the initial forecast at the start of training. $\lambda_{1}$ and $\lambda_{2}$ are both set to 0.5, inherited from the original DYffusion implementation. 

\item \textbf{Cold Sampling.} Clamped the cold sampling update to avoid passing the Interpolator data $\notin [0, 1]$. The cold sampling update from Algorithm 2, Step 4 in the original DYffusion paper becomes:
\begin{equation}
\hat{x}_{t+i_{n+1}} = \text{Clamp}_{[a,b]}(\mathrm{I}_{\phi}(x_{0}, \hat{x}_{t+h}, i_{\bf{n+1}}) - \mathrm{I}_{\phi}(x_{0}, \hat{x}_{t+h}, i_{\bf{n}}) + \hat{x}_{t+i_{n}})
\end{equation}
\end{enumerate}

\subsubsection{Baselines}\label{sec:baselines}
We compare DYffusion to auto-regressive implementations of both deterministic and statistical methods to align more closely with DYffusion's roll-out inference:
\begin{itemize}
\item \textbf{ConvLSTM} \cite{convlstm}: implemented the same 2-layer architecture used on the Radar Echo dataset, but with increased hidden states (128 vs 64), larger kernels (5$\times$5 vs 3$\times$3), and additional pixel-wise dropout (either 0.15 or 0.4) after each ConvLSTM cell for regularisation. The ConvLSTM was trained to predict the next timestep from the previous 4$\times$1$\times$128$\times$128 images (2 hours).
\item \textbf{Short-Term Ensemble Prediction System (STEPS)\cite{steps}}: implemented in PySteps \cite{pysteps} following the STEPS example on the PySTEPS website\footnote{\url{https://pysteps.readthedocs.io/en/stable/auto_examples/plot_steps_nowcast.html}}. The previous 4$\times$1$\times$128$\times$128 images (2 hours) were used as input and the units were transformed from $\text{mm} \cdot \text{h}^{-1}$ to dBR before estimating the motion field and forecasting the next timestep.
\end{itemize}

\section{Results}
\begin{table}[h]
\centering
\caption{The 4-hour averaged results on the test dataset for each of the evaluated models. The best metric score is highlighted in black and the second best score is highlighted in blue. The CRPS is calculated on the transformed data ($\in [0, 1]$), explaining the lower magnitude. The CRPS and SSR metrics for the deterministic ConvLSTM models are zeroed. CSI thresholds were evaluated at 2, 10 and 18 (in $\text{mm} \cdot \text{h}^{-1}$) to cover weak, moderate and heavy rain classifications respectively. The sampling times were recorded using an NVIDIA L4 GPU.}
\setlength{\tabcolsep}{4pt}
\begin{tabular}{lcccccccc}
\toprule
Model & $\downarrow$ MSE & $\downarrow$ LPIPS & $\uparrow$ CSI$_{2}$ & $\uparrow$ CSI$_{10}$ & $\uparrow$  CSI$_{18}$ & $\downarrow$ CRPS & $\uparrow$ SSR & Time [s] \\
\midrule
DYffusion$_{\text{LCB}}$ & \textcolor{blue}{\textbf{0.0020}} & \textbf{0.145} & \textbf{0.270} & \textbf{0.112} & \textbf{0.064} & \textcolor{blue}{\textbf{0.013}} & \textcolor{blue}{\textbf{0.131}} & 3 \\
DYffusion$_{\text{L1}}$ & \textbf{0.0015} & 0.323 & \textcolor{blue}{\textbf{0.227}} & \textcolor{blue}{\textbf{0.068}} & \textcolor{blue}{\textbf{0.023}} & \textbf{0.010} & 0.050 & 3 \\
ConvLSTM$_{\text{LCB}}$ & 0.0085 & \textcolor{blue}{\textbf{0.272}} & 0.139 & 0.027 & 0.010 & $-$ & $-$ & 1.2 \\
ConvLSTM$_{\text{BCE}}$ & 0.0052 & 0.335 & 0.165 & 0.038 & 0.016 & $-$ & $-$ & 1.2 \\
STEPS & 2.4102 & 0.342 & 0.140 & 0.030 & 0.011 & 0.013 & \textbf{0.219} & 2.1 \\
\bottomrule
\end{tabular}
\label{tab:model-comparison}
\end{table}

\begin{figure}
  \centering
  \includegraphics[width=0.7\linewidth]{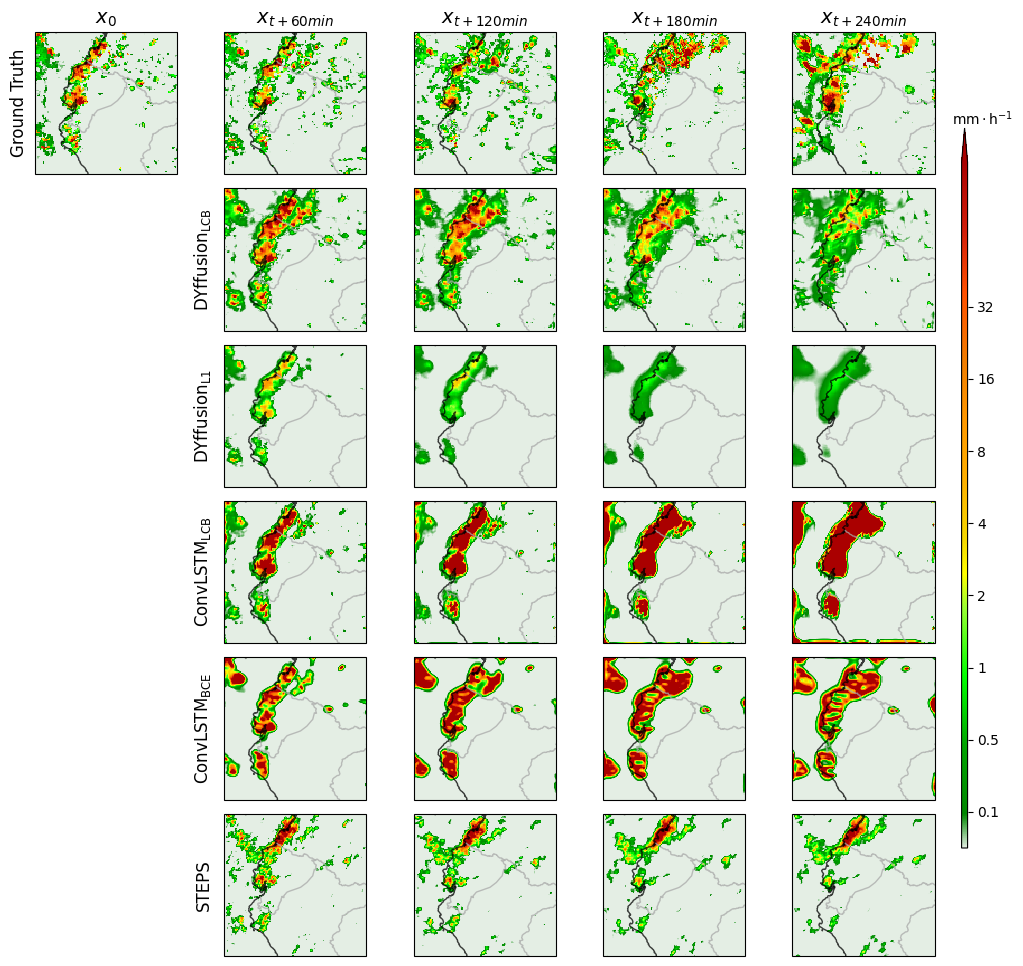}
  \caption{The evolution of Cyclone Yaku over 4 hours beginning at 03:00 UTC on March 9, 2023. The hourly forecast values (undersampled) are shown for each of the five evaluated models.}
  \label{fig:irp_plots}
\end{figure}

Table \ref{tab:model-comparison} shows the 4-hour averaged forecast evaluation metrics for each of the models on the test dataset. For both DYffusion and the baseline ConvLSTM \cite{convlstm}, two models were trained. One model was trained using the LCB loss and a second model was trained using the model's native loss function: L1 loss for DYffusion and binary cross-entropy (BCE) loss for ConvLSTM. The DYffusion and STEPS models are evaluated by taking the mean prediction from a 10-member ensemble. Figure \ref{fig:irp_plots} demonstrates each model's nowcasting ability on the heavy precipitation event, Cyclone Yaku \cite{nasa_peru_2023}. The forecasts clearly show the improved sharpness, detail and stability of DYffusion$_{\text{LCB}}$, especially up to a 2-hour horizon. This is supported by its outperforming CSI and LPIPS scores in Table \ref{tab:model-comparison}. DYffusion$_{\text{LCB}}$ can distinguish between heavy rain intensities in the band of rain moving toward the coast, and it tracks the split at the top of the rain-band forming at around $t+120$ min. The ConvLSTM$_{\text{LCB}}$ also demonstrates improved sharpness (at earlier timesteps) compared to the ConvLSTM$_{\text{BCE}}$, indicating the effectiveness of the LCB loss function at capturing the small-scale features. DYffusion$_{\text{LCB}}$ more accurately forecasts Cyclone Yaku's spatial expansion compared to STEPS. However, Figure \ref{fig:irp_plots} shows that optical flow methods remain effective at earlier timesteps, before chaotic behaviour emerges.

\paragraph{Computational Resources.} DYffusion was trained on an NVIDIA L4 GPU, which is readily accessible through cloud computing vendors at reasonable costs. Training took 20 minutes per epoch for the Interpolator (60 epochs, $\sim$20 hours total) and 1.5 hours per epoch for DYffusion with a 10-member ensemble (25-30 epochs, $\sim$40 hours total). The IMERG dataset requires 18GB of storage.

\section{Conclusion}
In this study, DYffusion has been extended to the task of precipitation nowcasting. The task was to forecast IMERG precipitation data at 30-minute intervals up to a 4-hour horizon, generating a total of 8$\times$1$\times$128$\times$128 images. By applying key modifications to the DYffusion framework and introducing a novel loss function, the modified framework outperforms both deterministic (ConvLSTM) and statistical (STEPS) baselines. Overall, the initial results of DYffusion$_{\text{LCB}}$ suggest that with some improvements, the DYffusion framework is a potential candidate for operational precipitation nowcasting. The constant memory footprint associated with the forecast-interpolate process makes it an attractive alternative to single-pass generative networks. 

\paragraph{Limitations.} Nowcasting precipitation is more challenging than SST or Navier-Stokes. The initial task of forecasting the 4-hour horizon from a single timestep is difficult. The implemented U-Net backbone struggles to not forecast $\hat{x}_{t+h}^{initial}$ as an expansion or reduction of the initial condition $x_{0}$. This ultimately constrains the problem to the spatial domain of $x_{0}$, contributing to the limited variability in the ensemble forecasts. 

\paragraph{Future Work.} The results of \cite{diffganensemble} demonstrate the importance of high SSR scores for more skilful precipitation nowcasts. Conditioning the initial forecast $\hat{x}_{t+h}^{initial}$ with atmospheric information, such as divergence at 925 hPa and wind speed, is an interesting avenue to explore. The additional physical context should help the model capture the non-linear chaotic evolution of the rainfall, especially at earlier timesteps, similar to the motion field in STEPS. This should improve $\hat{x}_{t+h}^{initial}$ and, therefore, the overall stochasticity by allowing the framework to sufficiently explore $P(x_{t+i_{1}:t+h} | x_{0})$.

\clearpage
\bibliographystyle{vancouver}
\bibliography{refs}

\end{document}